\documentclass[12pt]{amsart}
\usepackage[table]{xcolor}
\usepackage{times}
\usepackage{amssymb, amsmath}
\usepackage{amsthm}
\usepackage{tikz}
\usepackage{bm}
\usetikzlibrary{matrix,arrows}
\usepackage{enumitem}
\usepackage{booktabs}

\theoremstyle{plain}
\newtheorem{thm}{Theorem}

\theoremstyle{remark}
\newtheorem{remark}[thm]{Remark}
\newtheorem{example}[thm]{Example}

\newcommand{\bs}{\boldsymbol}

\begin{document}

\title[Variance decomposition in feature selection]{Orthogonal variance decomposition based feature selection}
\author[F. Kamalov]{Firuz Kamalov}
\address{Canadian University Dubai, Dubai, UAE.}
\email{\textcolor[rgb]{0.00,0.00,0.84}{firuz@cud.ac.ae}}
\keywords{feature selection; variance decomposition; Sobol decomposition; sensitivity index; total sensitivity index; wrapper methods}
\address{}
\email{}
\date{Mar 5, 2019}

\begin{abstract}
Existing feature selection methods fail to properly account for interactions between features when evaluating feature subsets. In this paper, we attempt to remedy this issue by using orthogonal variance decomposition to evaluate features. The orthogonality of the decomposition allows us to directly calculate the total contribution of a feature to the output variance. Thus we obtain an efficient algorithm for feature evaluation which takes into account interactions among features. Numerical experiments demonstrate that our method accurately identifies relevant features and improves the accuracy of numerical models.
\end{abstract}

\maketitle

\section{Introduction}
Feature selection is an important preprocessing step in machine learning tasks involving large datasets. Identifying and selecting the most relevant features enhances interpretability of the results and reduces computational cost. Feature selection methods can be broadly divided into three categories: filter methods, wrapper methods, and embedded methods. Filter methods use independent measures such as information gain or $\chi^2$-statistic to evaluate the importance of a feature subset. Wrapper methods evaluate a feature subset based on the accuracy of the learning model built with the given feature subset. Embedded methods perform feature evaluation as part of the model building process, with lasso regression being the canonical example of such approach. Among the three types of selection methods filter algorithms are the fastest to execute . However, most of  the existing filter methods ignore the interactions between feature variables. For instance, a basic filter method would rank features based on the information gain between an individual feature and the target variable.  However, it is possible that a pair of features, that individually have low information gain with respect to the target variable, may have a very high information gain if considered jointly. The canonical example where feature interactions are particularly prominent is the XOR problem.  

There have been various attempts in the literature to address the effects of feature interactions. The authors of the maximum relevance minimum redundancy method \cite{Peng} propose to incrementally add features to the optimal feature subset by maximizing the joint information gain between the feature subset and the target variable  while minimizing the average information gain between the pairs of feature variables. However, this method and other similar approaches take only a partial account of feature interactions and thus cannot guarantee an optimal solution. In addition, existing heuristics for exhaustive feature selection are time consuming. 

In this, paper we propose a feature selection method that under certain conditions allows for an efficient and complete feature evaluation. Our method is based on orthogonal decomposition of the variance of the model output \cite{Sal1, Sobol1}. In particular, the variance of the target variable $Y$ can be decomposed as 
\begin{equation}\label{eq_c}
V(Y) = \sum_i V_i + \sum_{i, j} V_{ij} + ... + V_{12 .. k}, 
\end{equation}
where each term $V_{i_1 i_2 .. i_s}$  represents the contribution - to the variance of $Y$ - stemming from feature interactions in subset $\{X_{i_1}, X_{i_2},... X_{i_s}\}$. Based on the variance decomposition we calculate the Total Sensitivity Index (TSI) of a feature which takes into account the interactions of a feature with all other features in determining its effect on the output variance. The TSI can thus be used to evaluate the importance of a feature with respect to the model output. Our approach is similar to wrapper methods as  it is trained on a particular learning model. However, it is more efficient than wrapper methods such as RFE in that the model being used needs to be trained only once. Thereafter all the remaining calculations are performed directly on the data using appropriate Monte Carlo integrals.

In Section 2, we give a brief overview of existing feature selection methods. In Section 3, we discuss decomposition of the output variance into its different components based on feature subsets. We give the necessary mathematical background and a Monte-Carlo method to perform required integral calculations. In Section 4, we present the results of numerical experiments that were performed using the proposed method. The results show that our method correctly selects relevant features and improves accuracy of learning models.

%%%%%%%%%%%%%%%%%%%%%%%%%%
\section{Related Work}
Feature selection algorithms can be broadly divided into three subsets: filter, wrapper and embedded methods. Filter methods use an independent metric - usually based on information gain \cite{Vergara} or $\chi^2$-statistic \cite{Jin} - to measure the importance of a feature subset. This is done by measuring the strength of the relationship between the features and the target class. The main advantage of filter methods is their relative robustness to changes in the learning model. In other words, a feature subset chosen with a filter method would perform similarly under different learning models. Wrapper methods evaluate the importance of a feature subset in the context of a particular model. The classical wrapper method uses a fixed classification (or regression) model to calculate its accuracy using a feature subset. The accuracy results determine the worth of the subset. The advantage of the wrapper method is that it is fine tuned to a particular model and thus gives better results than the filter approach on the particular chosen model. However, feature subsets that perform well on the original model do not perform as well on other models. In addition, wrapper methods are computationally intensive as the model must be trained for every subset evaluation. Embedded methods perform feature selection as part of the model building process as in the case of the lasso regression \cite{Tibshirani}.

The simplest approach to filter features is to evaluate the importance of each feature using a statistic that measures the relationship between a feature and the target variable and then select the features with the top scores. However, this approach does not take into account interactions between the features. It is quite possible that a pair of features that are individually irrelevant can have a high predictive power when joined together as in the case of XOR problem. In \cite{P1, Y}, the authors propose to evaluate features based on both their relevancy with respect to the target variable and their redundancy with respect to other features. In other words, the features that are highly correlated with the target variable but uncorrelated with other feature variables are given the preference. Authors use information gain as the measure of relevancy/redundancy though any other evaluation criteria would work as well. The authors in \cite{Kamalov} use information gain and $\chi^2$-statistic jointly to evaluate features. In particular, the two metrics  are combined into a single vector whose magnitude determines the importance of a feature. Combining the two metrics allows for lower variance in feature scores. This approach was shown to significantly reduce the number of features without hurting the accuracy of the model. In particular, the method was used effectively in selecting relevant factors in autism classification \cite{Thabtah}. 

There also exist filter methods that don't use the classical metrics from statistics or information theory. In \cite{Dash}, for instance, the authors propose to evaluate feature subsets based on the inconsistency measure whereby a subset is deemed inconsistent if there at least two instances with the same features labels that belong to different classes. The authors also discuss different search methods for finding the optimal subset based on the inconsistency measure.

Wrapper methods use specific models to evaluate features. One of the  popular wrapper algorithms is Recursive Feature Elimination (RFE). In this method, features are recursively dropped from the initial set based on their weights in the model. For instance, we can use OLS to first fit a hyperplane to the (normalized) data. Then we drop the feature with the lowest coefficient in the hyperplane equation. We again fit a hyperplane to our data but now using the reduced feature set. This procedure is repeated until some stopping criteria is achieved. RFE has implementations with many estimators including SVM  \cite{Guyon}, Random Forests \cite{Granitto}, and MLP \cite{Yang}. Although wrapper methods generally perform well their biggest drawback is high computational cost.

In \cite{Efimov}, the authors use Sobol's method \cite{Sobol1, Sobol2}  of decomposing output variance to evaluate features. In this approach, the variance of the target variable is deconstructed into a sum of variances based on feature subsets.  In particular, the authors use first order sensitivity index to evaluate each feature 
$$S_i = \frac{Var \big( E[Y|X_i] \big)}{V(Y)}.$$
In \cite{Kamalov2}, the authors use total sensitivity index based on variance decomposition to evaluate features. Total sensitivity index takes a more complete account of feature interactions and ultimately yields better results. In this paper, we pursue a similar approach and use total sensitivity analysis to evaluate features.

%%%%%%%%%%%%%%%%%%
\section{Orthogonal Variance Decomposition}
In this section, we will go through the steps of decomposing model output variance into  orthogonal components based on feature subsets. We will discuss how to use variance decomposition to evaluate features. In the end, we will describe a method for calculating the required integrals using a Monte-Carlo approach. Our presentation follows that of Saltelli \cite{Sal1} with further details provided in \cite{Homma, Sobol1, Sobol2}.

Suppose that target variable $Y$ is a function of a set of feature variables $\{X_1, X_2, ...X_k\}$, i.e., $Y = f(X_1, X_2, ...X_k)$.  Assume that the features $\{X_i\}$ are independently and uniformly distributed over the interval $[0, 1]$. We define 
\begin{equation}
\begin{split}
f_0 &= E[Y]\\
f_i(x) &= E[Y|X_i=x] - f_0\\
f_{i  j}(x,y) &= E[Y|X_i=x, X_j=y] - f_i(x) -f_j(y) -f_0
\end{split}
\end{equation}
and similarly for higher orders. For convenience, we drop the function arguments from the notation. Then it is not hard check that we obtain the following functional decomposition
\begin{equation}\label{eq_fun}
f = f_0 +\sum_i f_i + \sum_{i, j} f_{i j} + ... + f_{12 .. k} 
\end{equation}

\begin{example}
Let us demonstrate Eq. (\ref{eq_fun}) in the case of two features
\begin{equation*}
\begin{split}
f &= f_0 + f_1 + f_2 + f_{1 2}\\
&= f_0 + f_1 + f_2 + (E[Y|X_1, X_2] - f_1-f_2 -f_0)\\
&= E[Y|X_1, X_2]\\
&= Y
\end{split}
\end{equation*}
\end{example}

\begin{remark}
Our assumptions that the features are independent and uniformly distributed are highly optimistic. In practice, features have various non uniform distributions and are not independent of one another. In order to properly deal with these, more complex, scenarios one can employ other techniques for sensitivity analysis as described in \cite{Sal2}. However, as it will be demonstrated by our numerical experiments, even under the "naive" assumption of i.i.d. our feature selection algorithm performs very well. 
\end{remark}

The functional decomposition in Eq. (\ref{eq_fun})  is orthogonal in the sense that 
\begin{equation}\label{eq_orth}
\int_0^1 f_{i_1 i_2 .. i_s} \, dx_j = 0
\end{equation}
for every subset $\{X_{i_1}, X_{i_2},... X_{i_s}\}$ and $j\in \{i_1, i_2, .., i_s\}$. 

\begin{example}
Let us examine Eq. (\ref{eq_orth}) in the case of two features. Since $\int_0^1 E[Y|X_i] \, dx_i = E[Y]$, then we have
\begin{equation*}
\int_0^1 f_{1} \, dx_1 = \int_0^1 \big(E[Y|X_1] -f_0\big) \, dx_1 = 0.
\end{equation*}
Likewise, since $\int_0^1 E[Y|X_i, X_j] \, dx_i = E[Y|X_j]$ and $\int_0^1 E[Y|X_j] \, dx_i = E[Y|X_j]$, then
\begin{equation*}
\begin{split}
\int_0^1 f_{ij} \, dx_i &= \int_0^1 \big(E[Y|X_i, X_j] - f_i - f_j -f_0\big) \, dx_i \\
&=\int_0^1 \big(E[Y|X_i, X_j] - E[Y|X_i] + f_0 - E[Y|X_j] + f_0 -f_0\big) \, dx_i\\
&= \int_0^1 \big(E[Y|X_i, X_j] \big) \, dx_i - f_0 + f_0 - E[Y|X_j] = 0.
\end{split}
\end{equation*}

\end{example}

Another, useful property of the terms of the functional decomposition in Eq. (\ref{eq_fun}) is
\begin{equation}\label{eq_exp}
E[f_{i_1 i_2 .. i_s}] = 0,
\end{equation}
which follows directly from Eq. (\ref{eq_orth}). 

To obtain the variance decomposition, we square and integrate the two sides of Eq. (\ref{eq_fun}) 
$$\int_{[0, 1]^k} f^2\, d\boldsymbol{X} = \int_{[0, 1]^k} \big(f_0 +\sum_i f_i + \sum_{i, j} f_{i j} + ... + f_{1 2 .. k}\big)^2  \, d\boldsymbol{X}.$$

Using Eq. (\ref{eq_orth}) we can eliminate most of the cross multiplied terms on the right hand side. Then the final result can be written in the form

\begin{equation}\label{eq_funsqr}
\int_{[0, 1]^k} f^2\, d\boldsymbol{X} - f_0^2= \int_{[0, 1]^k} \big(\sum_i f_i^2 + \sum_{i, j} f_{i j}^2 + ... + f_{1 2 .. k}^2\big)  \, d\boldsymbol{X}.
\end{equation}

Note that the left hand side of the Eq. (\ref{eq_funsqr}) represents the variance of the model output $V(Y)$. Also note that by Eq. (\ref{eq_exp}) we get $Var(f_{i_1 i_2 .. i_s}) = \int f_{i_1 i_2 .. i_s}^2 \, d\boldsymbol{X}$. We now obtain Eq. (\ref{eq_c}) that was stated in the introduction of the paper
\begin{equation*}
V(Y) = \sum_i V_i + \sum_{i, j} V_{i j} + ... + V_{1 2 .. k}.
\end{equation*}
We can also view the variance decomposition in Eq. (\ref{eq_c}) from a slightly different angle:
\begin{equation}\label{eq_decomp2}
V(Y) = \sum_i V'_i + \sum_{i, j} V'_{i j} + ... + V'_{1 2 .. k}, 
\end{equation}
where $V'_i = Var(E[Y|X_i])$, $V'_{i j} = Var(E[Y|X_i, X_j])- V'_i - V'_j$, and similarly for higher orders. In the context of Equation \ref{eq_decomp2},  $V_i'$ represents the variance of $Y$ due solely to the feature $X_i$, $V_{i j}'$ represents the variance due to interaction between features $X_i$ and $X_j$, and similarly for higher orders.
\begin{example}
Let us show that Eq. (\ref{eq_c}) and Eq. (\ref{eq_decomp2}) are equivalent in the case of two features. First, it is easy to see that 
$$V_i = Var(E[Y|X_i] - f_0) = Var(E[Y|X_i])= V'_i.$$
Next,
\begin{equation*}
\begin{split}  
V_{i j}& = Var(E[Y|X_i, X_j] -f_i- f_j- f_0)\\
&=\int_0^1 \int_0^1 \big(E[Y|X_i, X_j] -f_i- f_j- f_0\big)^2 \, dx_i dx_j\\
&= \int_0^1 \int_0^1 \big(E[Y|X_i, X_j]^2  + f_i^2+ f_j^2 + f_0^2\big) \, dx_i dx_j\\
&-\int_0^1 \int_0^1 2E[Y|X_i, X_j] \big(f_i + f_j + f_0\big) \, dx_i dx_j\\
&= \int_0^1 \int_0^1 \big(E[Y|X_i, X_j]^2  - f_i^2- f_j^2 - f_0^2\big) \, dx_i dx_j\\
&= Var(E[Y|X_i, X_j])- V'_i - V'_j = V'_{i j}.
\end{split}
\end{equation*}
We used the fact that $\int f_k f_l \, d\boldsymbol{x}=0$ and $\int E[Y|X_i, X_j]  f_k \, d\boldsymbol{x}=f_k^2$ in our calculations.
\end{example}

Based on  Eq. (\ref{eq_decomp2}) we define the \textit{first order sensitivity} index of feature $X_i$ by
\begin{equation}
S_i = \frac{V'_i}{V(Y)},
\end{equation}
which measures the contribution of $X_i$ alone to the output variance. We can use $S_i$ as a simple tool to perform feature evaluation and selection. In fact, Efimov and Sulieman \cite{Efimov} used this approach to design their feature selection method. A more comprehensive metric to evaluate features would be the  \textit{total sensitivity} index defined as the sum of all variance terms of Eq. (\ref{eq_decomp2}) that contain contributions of feature $X_i$:
\begin{equation}\label{eq_sti}
S_{T_i} = \frac{V'_i + \sum_j V'_{ij} + \sum_{j,k} V'_{ijk} + ... + V'_{12...i..k}}{V(Y)}. 
\end{equation}
To simplify the expression in Eq. (\ref{eq_sti}) we can use the following useful identity: 
\begin{equation}\label{eq_compl} 
V'_i + \sum_j V'_{ij} + \sum_{j,k} V'_{ijk} + ... + V'_{12...i..k} = V(Y) - Var(E[Y|\bs{X}_{\sim i}]),
\end{equation}
where $\bs{X}_{\sim i}$ is the vector of all features except $X _i$. Let us illustrate the identity in Eq. (\ref{eq_compl}) with an example based on three features.

\begin{example}
Suppose that we have $Y= f(X_1, X_2, X_3)$. Let us verify that 
$$V'_1 + V'_{12} + V'_{13}+ V'_{123} = V(Y) - Var(E[Y|\bs{X}_{\sim 1}]).$$
Indeed, we know that $Var(E[Y|\bs{X}_{\sim 1}]) = V'_{23} + V'_2 +V'_3$. Since $V(Y)$ is the full sum of partial variances the result follows.
\end{example}

Using Eq. (\ref{eq_compl}) we can rewrite the definition of the total sensitivity index as 
\begin{equation}\label{eq_sti2}
S_{T_i} = 1 - \frac{Var(E[Y|\bs{X}_{\sim i}])}{V(Y)}, 
\end{equation}
Our final task is to calculate $Var(E[Y|\bs{X}_{\sim i}])$. There are exist various estimators for $Var(E[Y|\bs{X}_{\sim i}])$ \cite{Homma, Jansen, Sobol3}. In this paper, we choose to follow the approach of Homma and Saltelli \cite{Homma}. Let $\bs{A}$ and $\bs{B}$ be a pair of independent sampling matrices. Let $j$ and $i$ denote row and column indexes respectively. Define $\bs{A_B}^{(i)}$ to be matrix $\bs{A}$, where its $i$th column replaced with the $i$th column of $\bs{B}$. Then our estimator is
\begin{equation}\label{eq_estim}
Var(E[Y|\bs{X}_{\sim i}]) = \frac{1}{n}\sum_{j=1}^n f(\bs{A})_j f(\bs{A_B}^{(i)})_j -f_0^2 
\end{equation}

%%%%%%%%%%%%%%%%%%%%%%%%%%%%%%%%%%%%%%%%%%%%%%%%%%%%%%%%%%%%%%%%%%%%%%%%
\section{Numerical Experiments}
In this section, we will apply the total sensitivity ($S_{T_i}$) index to evaluate and select features. We will first test our approach on simulated data where the relevant features are known and show that our method correctly identifies the right features. Then we apply our method to real world datasets and show that our method can reduce the number of features used in a model without losing its accuracy. 

The details of the feature evaluation process using the total sensitivity index are given below
\\
\\
\indent \textbf{Algorithm}
\\
\indent \line(1,0){350}
\\
\begin{enumerate}[label=\arabic*., itemsep=2ex]
\item Let $\bs{D}$ be a dataset consisting of  $n$ rows, $k$ features, and a target variable. Train a model $f$ (SVR, RF, NN, etc) on  $\bs{D}$.

\item Discard the target variable so that $\bs{D}$ consists of only feature variables. Calculate the average $f_0  =  \frac{1}{n}\sum_{j=1}^n f(\bs{D})_j$ and total variance $ V(\hat{Y}) =  \frac{1}{n}\sum_{j=1}^n f(\bs{D})_j^2 -f_0^2$.

\item Shuffle the rows of $\bs{D}$ and split it into 2 halves: $\bs{A}$ and $\bs{B}$. 

\item For each feature $X_i$, create matrix $\bs{A_B}^{(i)}$ by replacing the $i$th column of $\bs{A}$ with the $i$th column of $\bs{B}$.  Then calculate the corresponding partial variance using Eq. (\ref{eq_estim}).

\item For each feature $X_i$,  calculate the corresponding index $S_{T_i}$ using Eq. (\ref{eq_sti2}).

\end{enumerate}

We begin our experiments with dataset generated using the Friedman function.
\begin{example}
In this example, we generate $20$ independent features $X_i$ using the uniform distribution over the interval $[0, 1]$ and the normal distribution with mean $0.5$ and  standard deviation $0.25$. We use the first 5 features to calculate the target variable according the Friedman function \cite{Breiman, Friedman} :
\begin{equation}\label{eq_friedman1}
y(\bs{x}) = 10  \sin(\pi x_0 \ x_1) + 20 (x_2 - 0.5) ^2 + 10 x_3 + 5 x_4 + \sigma \cdot N(0, 1),
\end{equation}
where $\sigma$ represents the amount of noise added to the data.
The remaining features are thus redundant. We begin by calculating the sensitivity index using the original Friedman function as described in Equation \ref{eq_friedman1}. As can be seen from Figure \ref{fig_friedman_pure}, both $S_i$ and $S_{T_i}$ based approaches perform well when the noise level is zero. In fact, even when the features are generated using the normal distribution - as opposed to the uniform distribution as required by the theory - the sensitivity indices of the relevant features are higher than that of the redundant features. When the noise level is increased to $\sigma =1$ the $S_{T_i}$ continues to classify  correctly  the relevant features while $S_i$ struggles with normally distributed features. When the noise level is increased further to $\sigma =2$  $S_{T_i}$ continues to outperform $S_i$ although it can no longer classify confidently the relevant features under the normal distribution.

\begin{figure}\label{fig_friedman_pure}
\includegraphics[scale=0.5]{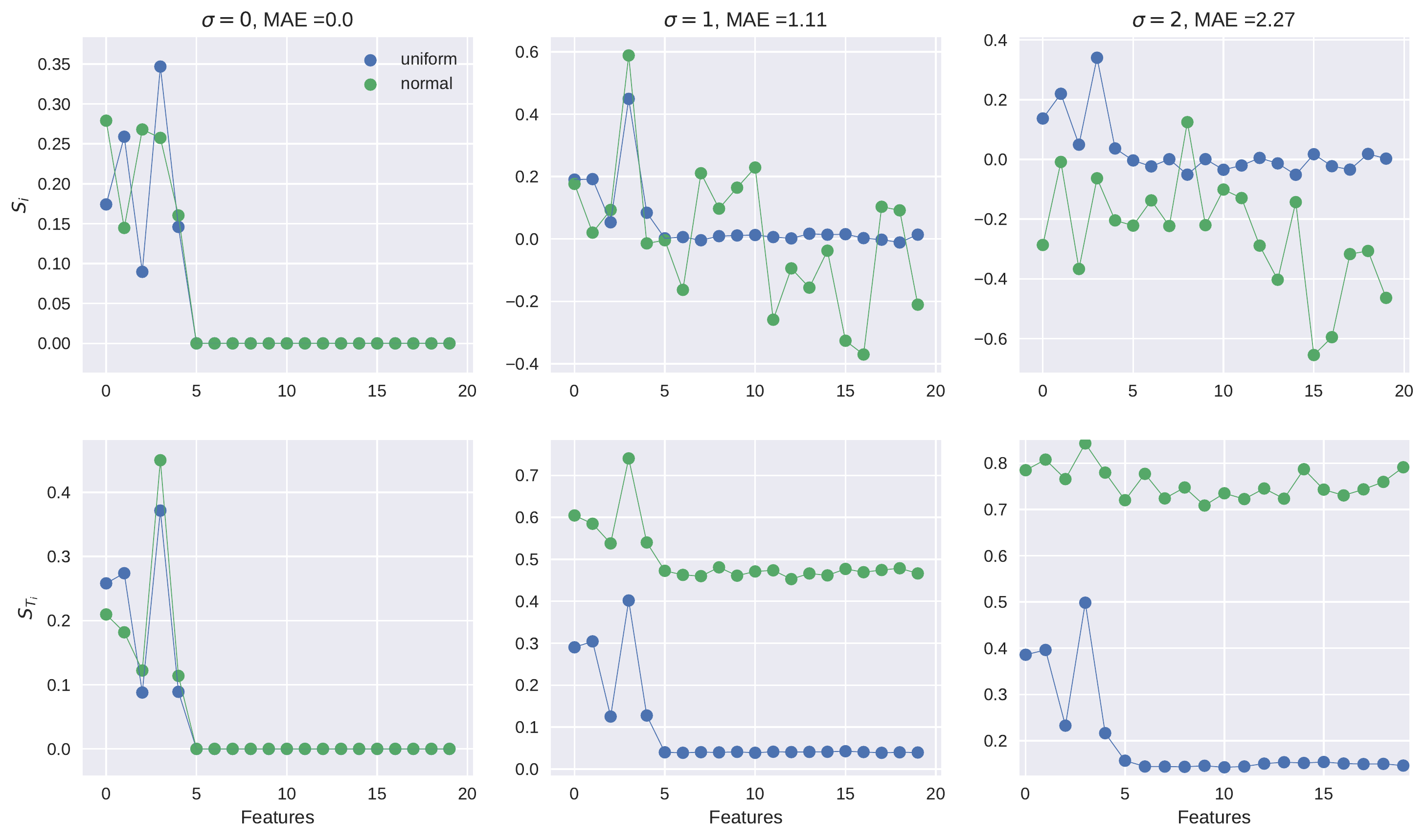}
\caption{$S_i$ and $S_{T_i}$ values based on the Friedman function under different levels of noise.}
\end{figure}

Next we train a Random Forrest (RF) regressor (with 10 estimators) \cite{Breiman2} on the data and use it to calculate the sensitivity index. As can be seen from Figure \ref{fig_friedman_rf}, the $S_{T_i}$ values are significantly higher for the relevant features. Somewhat surprisingly, the $S_{T_i}$ values are even higher for the normally distributed data which indicates that our method can perform well even when the theoretical assumptions on the data, i.e. i.i.d uniform distribution, are not strictly satisfied. We can also see from the last subplot that using $S_{T_i}$ we can identify the relevant features even under high levels of noise. 
\begin{figure}[h!]
\label{fig_friedman_rf}
\includegraphics[scale=0.5]{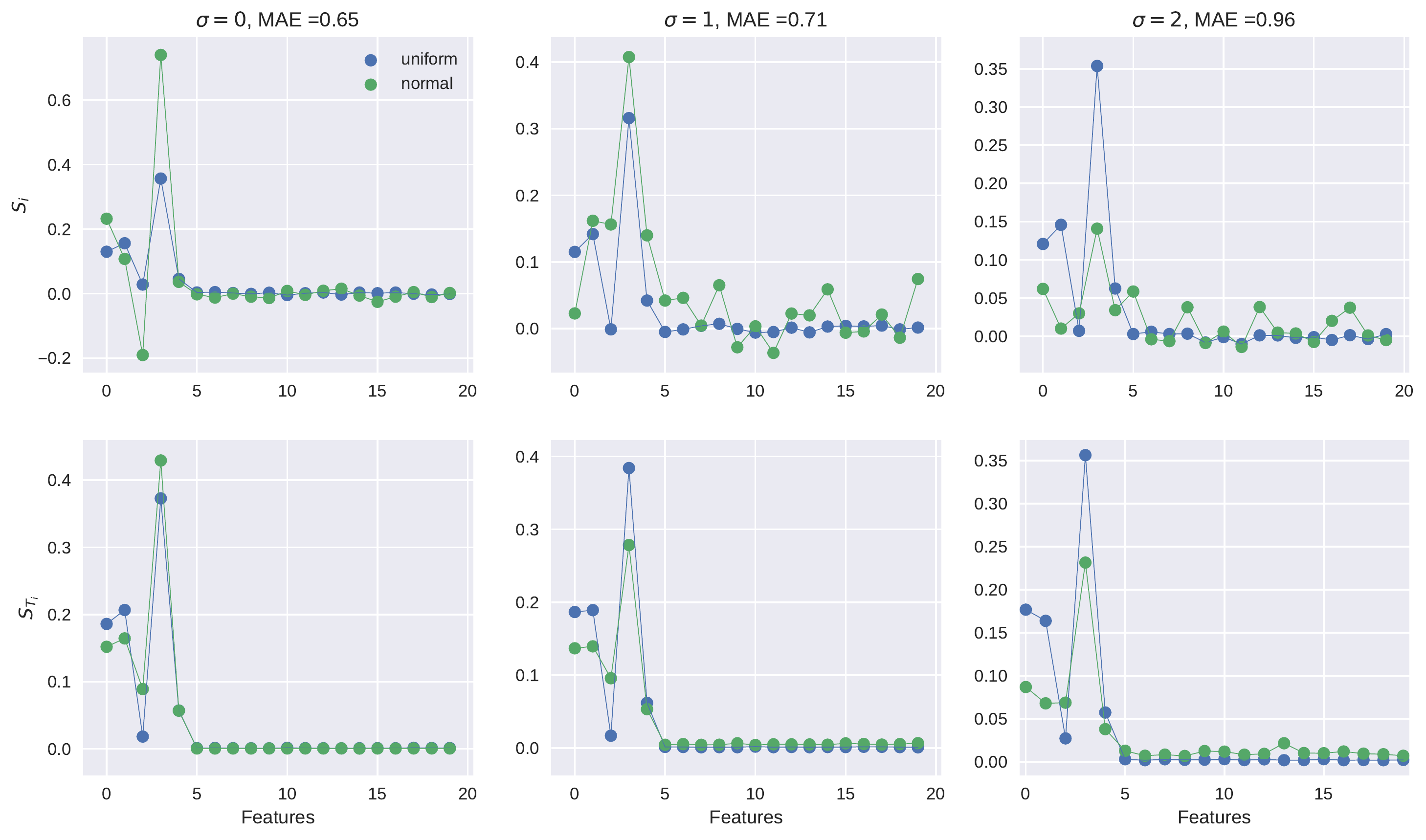}
\caption{$S_i$ and $S_{T_i}$ values based on the Random Forrest model under different levels of noise.}
\end{figure}

Finally, we use a trained neural network (NN) model \cite{Lecun} to calculate the sensitivity index. In our neural network model we used 2 hidden layers with 64 nodes in each layer. As shown in Figure \ref{fig_friedman_nn}, the $S_{T_i}$ values for uniformly distributed features are significantly higher than that of the redundant features.
Note that $S_{T_i}$ produces better results than $S_i$ as was the case in the previous scenarios.  However, when the noise level is increased the $S_{T_i}$ values for relevant features become very close to that of redundant features in the case of normally distributed data. Also note that the neural network model did not fit the data well as can be seen from the relatively high mean absolute error values. This might help to explain the fact that the neural networks approach underperformed relative to the previous two models.

\begin{figure}[h!]\label{fig_friedman_nn}
\includegraphics[scale=0.5]{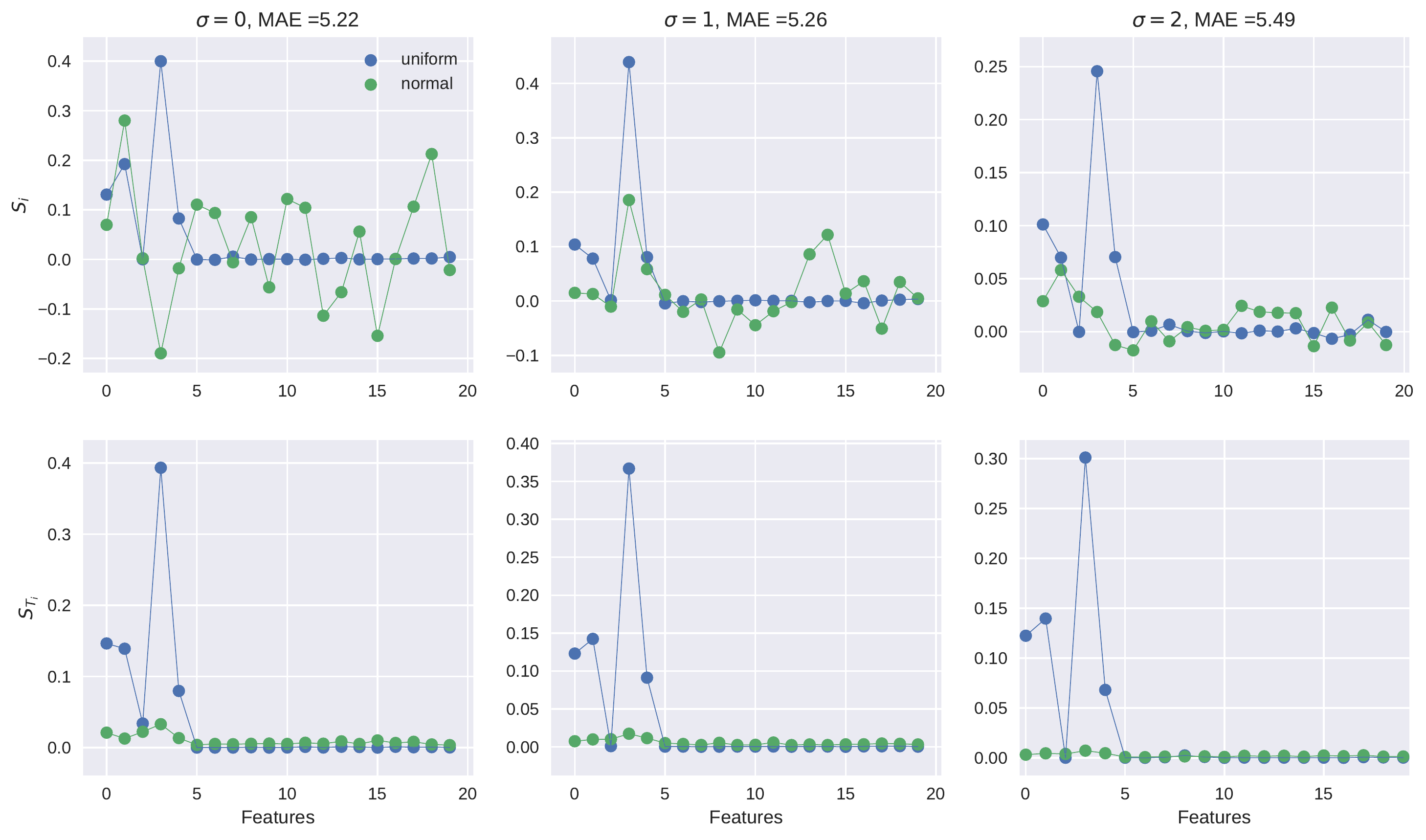}
\caption{$S_i$ and $S_{T_i}$ values based on the Neural Network model under different levels of noise.}
\end{figure}
\end{example}

\begin{example}
In this example, we generate a dataset consisting of $20$ independent features $X_i$  with the first four features taken as the relevant variables. The relevant features are uniformly distributed over the following intervals
\begin{itemize}
\item $0 \leq X_0 \leq 100$
\item $40 \pi \leq X_1 \leq 560 \pi$
\item $0 \leq X_2 \leq 1$
\item $1 \leq X_3 \leq 11.$
\end{itemize}
The remaining features are uniformly distributed over the interval $[0, 1]$.
The target variable is calculated according another Friedman function \cite{Breiman, Friedman} :
\begin{equation}\label{eq_friedman2}
y(\bs{x}) = \sqrt{x_0^2 + (x_1 x_2  - \frac{1}{x_1 x_3})^2}  + \sigma \cdot N(0, 1).
\end{equation}
We apply various feature ranking methods to test if they can correctly identify the relevant features. To this end, we train Support Vector Machines (SVR) \cite{Smola}, RF, and NN models and use them to calculate the sensitivity indices. 
We also use the original function from Equation \ref{eq_friedman2} to calculate the corresponding sensitivity indices.
We benchmark the performance of the $S_{T_i}$ based approach to the performance of RFE methods. In particular, we use the RFE algorithms based on SVR, Linear Regression (LR), and RF regressors.  
The results  in Table \ref{tb_Friedman2} show that the $S_{T_i}$ based approach produces the best results in correctly identifying the relevant features. In particular,  $S_{T_i} Friedman$ and  $S_{T_i} SVR$ rank the relevant features in top 4 even when we apply noise to the data. Similarly, $S_{T_i} RF$ and  $S_{T_i} NN$ perform well with only a single feature (4) being ranked one spot below the top 4. By comparison RFE $SVR$ and RFE $LR$ rank only one feature correctly. We note that RFE $RF$ does perform well, somewhat surprisingly,  when noise is applied to the data.

\begin{table}[h!]\label{tb_Friedman2}
\caption{Rankings of the four relevant features (1, 2, 3, and 4) using various feature selection methods. The dataset used consists of 20 features with the target variable being generated via the Friedman function (Equation \ref{eq_friedman2}). }

\centering

\begin{tabular}{@{} lllll | llll @{}}
\toprule
Noise & \multicolumn{4}{c}{$\sigma = 0$} & \multicolumn{4}{c}{$\sigma = 10$}\\
\toprule
Feature & 1 & 2 & 3 & 4 & 1 & 2 & 3 & 4\\ 
\midrule

\textit{$S_i$ Friedman} & 20&   2&   1&  19&   19&  2&  1&  5\\

\textit{$S_{T_i}$Friedman} & 3&   2&   1&   4&  3&  2&  1&  4 \\

\textit{$S_i$ SVR} & 20&  1&  2&  3&   20&  1&  2&  3 \\

\textit{$S_{T_i}$SVR} & 2&  1&  4&  3&   2&  1&  4&  3 \\

\textit{$S_i$ RF} &   13&  2&  1& 19&  20&  2&  1& 11\\

\textit{$S_{T_i}$RF}&  3&  2&  1&  5&  3&  2&  1&  5\\

\textit{$S_i$ NN} &   2&  1& 20&  4 &  19&  1& 20&  9\\ 

\textit{$S_{T_i}$NN} &  2&  1&  3& 1&  2&  1&  3&  5\\ 

\textit{RFE SVR} & 19& 17&  1& 14&  19& 17&  1& 11\\ 

\textit{RFE LR} & 17&  18&  1& 19& 18& 19&  1& 20 \\

\textit{RFE RF}& 6  &2  &1  &7 &3  &2  &1  &6\\ \bottomrule 
\end{tabular}
\end{table}
\end{example}

\begin{example}
In this example, we generate a dataset consisting of $20$ independent features $X_i$  with the first four features taken as the relevant variables. The relevant features are uniformly distributed over the following intervals
\begin{itemize}
\item $0 \leq X_0 \leq 100$
\item $40 \pi \leq X_1 \leq 560 \pi$
\item $0 \leq X_2 \leq 1$
\item $1 \leq X_3 \leq 11.$
\end{itemize}
The remaining features are uniformly distributed over the interval $[0, 1]$.
The target variable is calculated according another Friedman function \cite{Breiman, Friedman} :
\begin{equation}\label{eq_friedman2}
y(\bs{x}) = \sqrt{x_0^2 + (x_1 x_2  - \frac{1}{x_1 x_3})^2}  + \sigma \cdot N(0, 1).
\end{equation}
We apply various feature ranking methods to test if they can correctly identify the relevant features. To this end, we train Support Vector Machines (SVR) \cite{Smola}, RF, and NN models and use them to calculate the sensitivity indices. 
We also use the original function from Equation \ref{eq_friedman2} to calculate the corresponding sensitivity indices.
We benchmark the performance of the $S_{T_i}$ based approach to the performance of RFE methods. In particular, we use the RFE algorithms based on SVR, Linear Regression (LR), and RF regressors.  
The results  in Table \ref{tb_Friedman2} show that the $S_{T_i}$ based approach produces the best results in correctly identifying the relevant features. In particular,  $S_{T_i} Friedman$ and  $S_{T_i} SVR$ rank the relevant features in top 4 even when we apply noise to the data. Similarly, $S_{T_i} RF$ and  $S_{T_i} NN$ perform well with only a single feature (4) being ranked one spot below the top 4. By comparison RFE $SVR$ and RFE $LR$ rank only one feature correctly. We note that RFE $RF$ does perform well, somewhat surprisingly,  when noise is applied to the data.

\begin{table}\label{Friedman3}
\caption{Rankings of the four relevant features (1, 2, 3, and 4) using various feature selection methods. The dataset used consists of 20 features with the target variable being generated via the Friedman function (Equation \ref{eq_friedman2}). }
\begin{tabular}{@{} lllll | llll @{}}
\toprule

Noise & \multicolumn{4}{c}{$\sigma = 0$} & \multicolumn{4}{c}{$\sigma = 10$}\\
\toprule

Feature & 1 & 2 & 3 & 4 & 1 & 2 & 3 & 4\\ 
\midrule

\textit{FSI Friedman} &  2&   3&   1&  20& 13&  5&  9&  1\\

\textit{TSI Friedman} & 3&   2&   1&   4&  15&  4&  6&  1\\

\textit{FSI SVR} &20&  1&  2& 19&  3&  1&  7&  2 \\

\textit{TSI SVR} & 2&  1&  4&  3&  2&  1&  5&  3\\

\textit{FSI RF} &   3&  2&  1& 10&  9&  8&  1& 14\\

\textit{TSI RF}&  3&  2&  1& 16& 4&  2&  1& 10\\

\textit{FSI NN} &   4&  1& 20&  3& 2&  1&  8&  3 \\ 

\textit{TSI NN} &  2&  1&  4&  3 &  2&  1&  4&  3\\ 

\textit{RFE SVR} & 19&  20&  1& 16& 19& 20&  1& 17\\ 

\textit{RFE LR} & 18 &20  &6 &16  &17 &20 & 1 &15\\ 

\textit{RFE RF}& 10&  8&  1& 12&  9&  2&  1& 20\\ \bottomrule 
\end{tabular}
\end{table}

\end{example}

\section{Conclusion}
In this paper we discuss a new approach to evaluating features based on total sensitivity index (TSI). There are two main advantages to using TSI in feature evaluation. First, the TSI incorporates the effects of interactions between the features. Most of the modern feature selection methods do not fully consider the effects that other features have on the relationship between a feature and the target class. In this sense, TSI approach stands in a small crowd. Second advantage of TSI lies in its relative efficiency. Although it is not as efficient as a filter method, it is still much faster than most of other wrapper methods. 

The experiments with artificially generated data (Friedman data set) where the relevant features were known showed that TSI is the highest for the relevant features. In other, words TSI effectively identified the important features in the data set. In particular, TSI computed using the neural networks model can effectively identify the relevant features even with high level of noise in the data. We also tested our approach to a real life data set (Communities and Crime). The experiments with this data set showed that TSI is very competitive with other modern feature selection models. In particular, the highest performance on the data set was achieved using the features selected via TSI-SVR.
 
%%%%%%%%%%%%%%%%%%%%%%%%%%%%%%%%%%%%

%% References


%%%%%%%%%%%%%%%%%%%%%%%%%%%%%%%

\end{document}